\documentclass[nohyperref]{article}

\usepackage{microtype}
\usepackage{graphicx}
\usepackage{subfigure}
\usepackage{booktabs} 

\usepackage{hyperref}

\usepackage{amsmath}
\usepackage{amssymb}
\usepackage{mathtools}
\usepackage{amsthm}

\usepackage[capitalize]{cleveref}
\Crefname{section}{Sec}{Secs.}
\Crefname{figure}{Fig}{Figs.}
\Crefname{theorem}{Thm}{Thms.}
\Crefname{table}{Table}{Tables.}
\Crefname{appendix}{Supp}{Supp}

\theoremstyle{plain}

\theoremstyle{definition}

\theoremstyle{remark}

\usepackage[textsize=tiny]{todonotes}
\usepackage{fontawesome}
\usepackage{placeins}

\definecolor{cblue}{RGB}{8, 85, 153}
\usepackage[accepted]{icml2022}

\newcommand{\method}{Group Probability-Weighted Tree Sums}
\newcommand{\methodabbrv}{G-FIGS}

\icmltitlerunning{Group Probability-Weighted Tree Sums for Interpretable Modeling}

\begin{document}

\twocolumn[
\icmltitle{Group Probability-Weighted Tree Sums\\for Interpretable Modeling of Heterogeneous Data}
\icmlsetsymbol{equal}{*}

\begin{icmlauthorlist}
\icmlauthor{Keyan Nasseri}{eecs}
\icmlauthor{Chandan Singh}{eecs}
\icmlauthor{James Duncan}{biostat}
\icmlauthor{Aaron Kornblith}{ucsf,ucsf2}
\icmlauthor{Bin Yu}{eecs,biostat,stat}
\end{icmlauthorlist}

\icmlaffiliation{eecs}{EECS Department, UC Berkeley}
\icmlaffiliation{stat}{Statistics Department, UC Berkeley}
\icmlaffiliation{biostat}{Group in Biostatistics, UC Berkeley}
\icmlaffiliation{ucsf}{Emergency Medicine, UC San Francisco}
\icmlaffiliation{ucsf2}{Pediatrics Department, UC San Francisco}

\icmlcorrespondingauthor{Bin Yu}{binyu@berkeley.edu}

\icmlkeywords{Machine Learning, ICML}

\vskip 0.3in
]

\printAffiliationsAndNotice{}  

\begin{abstract}
Machine learning in high-stakes domains, such as healthcare, faces two critical challenges: (1) generalizing to diverse data distributions given limited training data while (2) maintaining interpretability. To address these challenges, we propose an instance-weighted tree-sum method that effectively pools data across diverse groups to output a concise, rule-based model. Given distinct groups of instances in a dataset (e.g., medical patients grouped by age or treatment site), our method first estimates group membership probabilities for each instance. Then, it uses these estimates as instance weights in FIGS
~\cite{tan2022fast}, to grow a set of decision trees whose values sum to the final prediction.
We call this new method \method~(\methodabbrv).
\methodabbrv~achieves state-of-the-art prediction performance on important clinical datasets; 
e.g., holding the level of sensitivity fixed at 92\%, \methodabbrv~increases specificity for identifying cervical spine injury (CSI) by up to 10\% over CART and up to 3\% over FIGS alone, with larger gains at higher sensitivity levels.
By keeping the total number of rules below 16 in FIGS,
the final models remain interpretable,
and we find that their rules match medical domain expertise.
All code, data, and models are released on Github.
\footnote{\method~is integrated into the imodels package \href{https://github.com/csinva/imodels}{\faGithub\,csinva/imodels}~\cite{singh2021imodels} with an sklearn-compatible API. Experiments for reproducing the results here can be found at \href{https://github.com/Yu-Group/imodels-experiments}{\faGithub\,Yu-Group/imodels-experiments}.}
\end{abstract}

\section{Introduction}

Recent advances in machine learning (ML) have led to impressive increases in predictive performance.
However, ML has high stakes in the healthcare domain, with two critical challenges to effective adoption.

First, models must adapt to heterogenous data from diverse groups of patients~\cite{ginsburg2018precision}.
Groups may differ dramatically and require distinct features for high predictive performance on the same outcome;
e.g., infants may be nonverbal, excluding features that require a verbal response, which in turn may be highly predictive in adults.
A potential solution is to simply fit a unique model to each group (e.g., \citeauthor{kuppermann2009identification}~\citeyear{kuppermann2009identification}), but this discards valuable information that can be shared across groups.

Second, a lack of interpretability is unacceptable in healthcare and many other domains~\cite{murdoch2019definitions,rudin2019stop}.
Interpretability is required to ensure that models behave reasonably, identify when models will make errors, and make the models amenable to inspection by domain experts.
Moreover, interpretable models tend to be much more computationally efficient than larger black-box models, often making them easier to use with humans in the loop, such as in medical diagnosis.

\begin{figure*}[h]
    \centering
    \includegraphics[width=.95\textwidth]{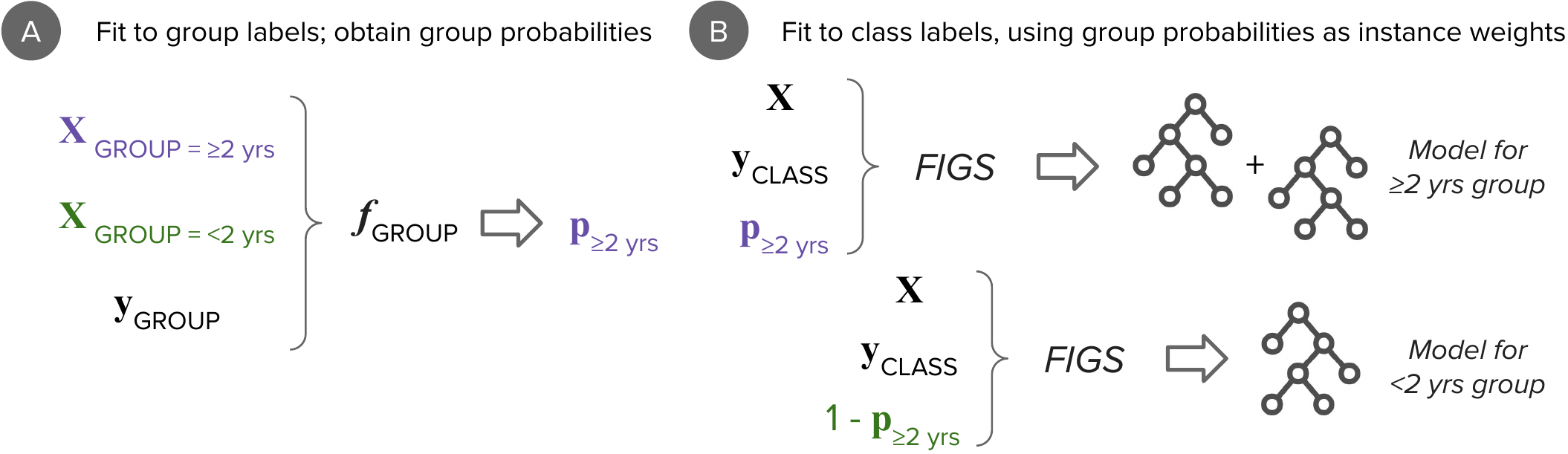}
    \vspace{-8pt}
    \caption{Overview of \methodabbrv. \textbf{(A)} First, the covariates of each instance in a dataset are used to estimate an instance-specific probability of membership in each of the pre-specified groups in the data (e.g., patients of age ${<}2$ \textit{yrs} and ${\ge}2$ \textit{yrs}). \textbf{(B)} Next, these membership probabilities are used as instance weights when fitting an interpretable model for each group.}
    \vspace{-8pt}
    \label{fig:intro}
\end{figure*}

 Here, we (1) address the challenge of sensibly sharing data
across groups using group membership probability estimates and (2) address the challenges of interpretability by outputting a concise rule-based model.
Specifically, we introduce \method~(\methodabbrv\footnote{Our method is abbreviated as \methodabbrv~because we use an instance-weighted version of Fast Interpretable Greedy-tree sums (FIGS, \citeauthor{tan2022fast}~\citeyear{tan2022fast}) to output a rule-based model.}), a two-step algorithm which takes in training data divided into known
groups (e.g., patients in distinct age ranges), and outputs a rule-based model (\cref{fig:intro}).
\methodabbrv~first fits a classifier to predict group membership probabilities for each input instance~(\cref{fig:intro}A).
 Next, it uses these estimates
as soft instance weights in the loss function of FIGS.
The output is an ensemble of decision trees where the contribution from each tree is summed
to yield a final prediction.

By sharing data sensibly across groups during training, \methodabbrv~results in a separate highly accurate rule-based model for each group.
We test \methodabbrv~on three real-world clinical datasets (\cref{sec:results}) and for two age groups commonly used in ER medicine; we find that \methodabbrv~outperforms state-of-the-art clinical decision instruments and competing ML methods in terms of specificity achieved at the high levels of sensitivity required in many clinical contexts.
Moreover, \methodabbrv~maintains interpretability and ease-of-vetting with small (1-3 trees per group) and concise ($\le 6$ splits per tree) clinical decision instruments by limiting the total number of rules across the trees for a given group.

\section{Background and related work}

We study the problem of sharing data across diverse groups in a supervised setting.
Our methodology relies on estimates of group membership probabilities as instance weights in each group's outcome model, selected via cross-validation among multiple probability estimation methods.
More weight is placed on instances that have higher estimated group-specific membership probability.
In their role as group-balancing weights, we use these probabilities in a manner that is mathematically (though not conceptually) analogous to the use of propensity scores in causal inference for adjusting treatment-effect estimates~\cite{guo2014propensity}.
More generally, this work is related to the literature on
transfer learning~\cite{zhuang2020comprehensive}, but we focus on transfer
in the setting where outcomes are known for all training instances and interpretability is crucial.

Intrinsically interpretable methods, such as decision trees, have had success as highly predictive and interpretable models~\cite{quinlan1986induction,breiman1984classification}.
Recent works have focused on improving the predictive performance of intrinsically interpretable methods~\cite{ustun2016supersparse,ha2021adaptive}, particularly for rule-based models~\cite{friedman2008predictive,agarwal2022hierarchical,tan2022fast,lin2020generalized}, without degrading interpretability.

A key domain problem involving interpretable models is the development of clinical decision instruments, which can assist clinicians in improving the accuracy and efficiency of diagnostic strategies.
Recent works have developed and validated clinical decision instruments using interpretable ML models, particularly in emergency medicine~\cite{bertsimas2019prediction,stiell2001canadian,kornblith2022predictability,holmes2002identification}.

\section{Method: \methodabbrv}


\paragraph{Setup.}
We assume a supervised learning setting (classification or regression) with features $X$ (e.g., \textit{blood pressure}, \textit{signs of vomiting}), and an outcome $Y$ (e.g., \textit{cervical spine injury}).
We are also given a group label $G$, which is specified using the context of the problem and domain knowledge; for example, $G$ may correspond to different sites at which data is collected, different demographic groups which are known to require different predictive models, or data before/after a key temporal event.
$G$ should be discrete, as \methodabbrv~will produce a separate model for each unique value of $G$, but may be a discretized continuous or count feature.

\paragraph{Fitting group membership probabilities.}
The first stage of \methodabbrv~fits a classifier to predict group membership probabilities $P(G|X)$ (\cref{fig:intro}A).\footnote{In estimating $P(G=g|X)$, we exclude features that trivially identify $G$ (e.g., we exclude \textit{age} when values of $G$ are age ranges).}
Intuitively, these probabilities inform the degree to which a given instance is representative of a particular group; the larger the group membership probability, the more the instances should contribute to the model for that group.
Any classifier can be used; we find that logistic regression and gradient-boosted decision trees perform best.
The group membership probability classifier can be selected 
using cross-validation, either via group-label classification metrics or downstream performance of the weighted prediction model; we take the latter approach.

\paragraph{Fitting group probability-weighted FIGS.}
In the second stage (\cref{fig:intro}B), for each group $G=g$, \methodabbrv~uses the estimated group membership probabilities, $P(G=g|X)$, as instance weights in the loss function of a ML model for each group $P(Y|X, G=g)$.
Intuitively, this allows the outcome model for each group to use information from out-of-group instances when their covariates are sufficiently similar.
While the choice of outcome model 
is flexible, 
we find that the Fast Interpretable Greedy-Tree Sums (FIGS) model~\cite{tan2022fast} performs best when both interpretability and high predictive performance are required.\footnote{When interpretability is not critical, the same weighting procedure could also be applied to black-box models, such as Random Forest~\cite{breiman2001random}.}
By greedily fitting a sum of trees, FIGS effectively allocates a small budget of rules to different types of structure in data.

\section{Results and discussion}
\label{sec:results}
\paragraph{Datasets and data cleaning.}

\cref{tab:datasets} shows the main datasets under consideration here.
They each constitute a large-scale multi-site data aggregation by the Pediatric Emergency Care Applied Research Network,
with a relevant clinical outcome.
For each of these datasets, we use their natural grouping of patients into ${<}2$ \textit{yrs} and ${\ge}2$ \textit{yrs} groups, where the young group includes only patients whose age is less than two years.
This age-based threshold is commonly used for emergency-based diagnostic strategies (e.g., \citeauthor{kuppermann2009identification}~\citeyear{kuppermann2009identification}), because it follows a natural stage of development, including a child’s ability to participate in their care.
At the same time, the natural variability in early childhood development also creates opportunities to share information across this threshold.
These datasets are non-standard for ML; as such, we spend considerable time cleaning and preprocessing these features along with medical expertise included in the authorship team.
\footnote{Details, along with the openly released clean data can be found in \cref{sec:data_preprocessing}.
Additionally, simulation results showing the effectiveness of \methodabbrv~are shown in \cref{sec:simulations}.}
We use 60\% of the data for training, 20\% for tuning hyperparameters (including estimation of $P(G|X)$), and 20\% for evaluating test performance of the final models.

\begin{table}[h]
    \centering
    \footnotesize
    \begin{tabular}{lrrrr}
\toprule
Name &  Patients &  Outcome &  \% Outcome &  Features \\
\midrule
 TBI &     42428 &      376 &        0.9 &        61 \\
 IAI &     12044 &      203 &        1.7 &        21 \\
 CSI &      3313 &      540 &       16.3 &        34 \\
\bottomrule
\end{tabular}
    \caption{Clinical decision-instrument datasets for traumatic brain injury (TBI)~\cite{kuppermann2009identification}, intra-abdominal injury (IAI)~\cite{holmes2002identification}, and cervical spine injury (CSI)~\cite{leonard2019cervical}.}
    \label{tab:datasets}
\end{table}

\vspace{-10pt}

\paragraph{\methodabbrv~predicts well.}

\cref{tab:results} shows the prediction performance of \methodabbrv~and a subset of baseline methods.
Sensitivity is extremely important for these settings,
as a false negative (missing a diagnosis) has much more severe consequences than a false positive.
For high levels of sensitivity, \methodabbrv~generally improves the model's specificity against the baselines.
We compare to three baselines: CART~\cite{breiman1984classification}, FIGS~\cite{tan2022fast}, and Tree-Alternating Optimization TAO~\cite{carreira2018alternating}).
For each baseline, we either (i) fit one model to all the training data or (ii) fit a separate model to each group (denoted with \textit{-SEP}).
Limits on the total number of rules for each model are varied over a range which yields interpretable models, from 2 to 16 maximum rules (full details of this and other hyperparameters are in \cref{sec:hyperparams}).

\begin{figure}[h]
    \centering
    \includegraphics[width=\columnwidth]{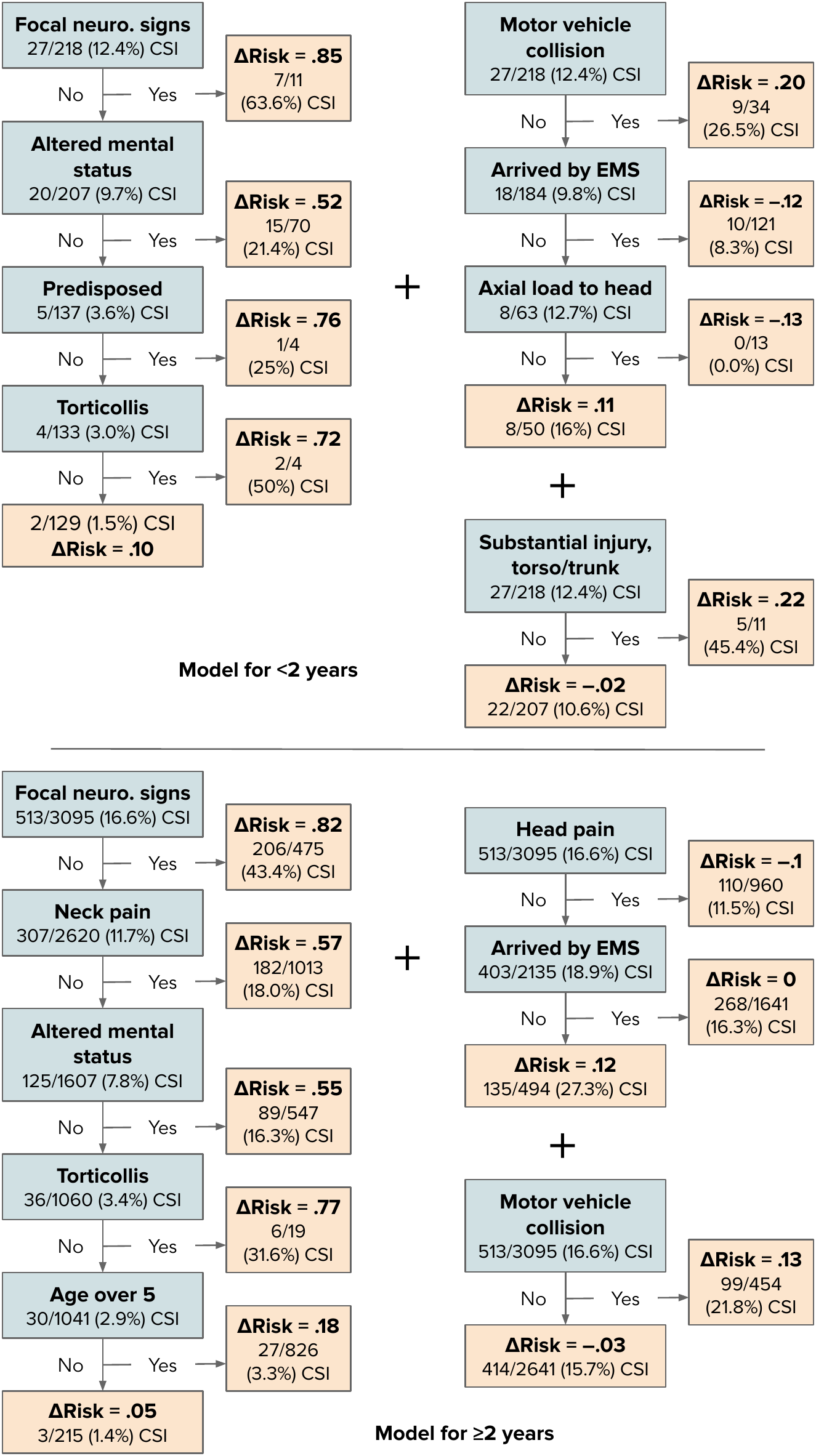}
    \vspace{-15pt}
    \caption{\methodabbrv~models fitted to the CSI dataset are concise, highly predictive, and match known medical knowledge.
    The left tree for ${<}2$ \textit{yrs} has high sensitivity (99\%); adding the upper right tree boosts specificity by 8.7\% and decreases sensitivity by 0.4\%.
    }
    \label{fig:model_ex_csi}
\end{figure}

\begin{table*}[h]
    \centering
    \footnotesize
        \begin{tabular}{lrrrrrrrrrrrr}
    \toprule
    {} & \multicolumn{4}{c}{Traumatic brain injury} &
        \multicolumn{4}{c}{Cervical spine injury} 
        & \multicolumn{4}{c}{Intra-abdominal injury} \\
     \cmidrule(lr){2-5}
     \cmidrule(lr){6-9}
     \cmidrule(lr){10-13}
     Sensitivity level: & 92\% & 94\% & 96\% & 98\% & 92\% & 94\% & 96\% & 98\% & 92\% & 94\% & 96\% & 98\% \\
     \midrule
    TAO & 6.2 & 6.2 & 0.4 & 0.4
        & 41.5 & 21.2 & 0.2 & 0.2 
        & 0.2  & 0.2 & 0.0 & 0.0 \\
     TAO-SEP & 26.7 & 13.9  & 10.4 & 2.4 
        & 32.5  & 7.0  & 5.4 & 2.5 
        & 12.1  & 8.5 & 2.0 & 0.0 \\
    CART & 20.9 & 14.8 & 7.8. & 2.1 
        & 38.6 & 13.7 & 1.5 & 1.1
        & 11.8 & 2.7 & 1.6 & 1.4 \\
     CART-SEP & 26.6 & 13.8 & 10.3 &  2.4 
        & 32.1 & 7.8 & 5.4 & 2.5 
        & 11.0 & 9.3 & 2.8 & 0.0 \\
    G-CART & 15.5 &  13.5 & 6.4 & 3.0
        &  38.5 &  15.2 & 4.9 & 3.9 
        & 11.7 &  10.1 & 3.8 &  0.7 \\
     FIGS & 23.8 &  18.2 &  12.1 &  0.4 
        & 39.1 & 33.8 & 24.2 & \textbf{16.7}   
        & \textbf{32.1} & 13.7 & 1.4 & 0.0  \\
     FIGS-SEP & 39.9 & 19.7&  \textbf{17.5} &  2.6 
        & 38.7 & 33.1 & 20.1 & 3.9 
        & 18.8 & 9.2 & 2.6 &  0.9  \\
     \textbf{\methodabbrv} & \textbf{42.0} & \textbf{23.0} & 14.7 & \textbf{6.4} 
        & \textbf{42.2}  & \textbf{36.2} & \textbf{28.4} & 15.7 
        & 29.7 & \textbf{18.8} & \textbf{11.7} & \textbf{3.0} \\
    \bottomrule
    \end{tabular}

        \vspace{-5pt}
        \caption{Best test set specificity when sensitivity is constrained to be above a given threshold. \methodabbrv~provides the best performance overall in the high-sensitivity regime.
        \textit{-SEP} models fit a separate model to each group, and generally outperform fitting a model to the entire dataset.
        G-CART follows the same approach as \methodabbrv~but uses weighted CART instead of FIGS for each final group model.
        Averaged over 10 random data splits into training, validation, and test sets, with hyperparameters chosen independently for each split.
        }
    \label{tab:results}
\end{table*}

\begin{table*}[h]
    \centering
    \footnotesize
    \begin{tabular}{lrlrlr}
\\
    \toprule
    \multicolumn{2}{c}{Traumatic brain injury} 
        & \multicolumn{2}{c}{Cervical spine injury} 
        & \multicolumn{2}{c}{Intra-abdominal injury}\\
     \cmidrule(lr){1-2}
     \cmidrule(lr){3-4}
     \cmidrule(lr){5-6}
     Variable & Coefficient & Variable & Coefficient & Variable & Coefficient\\
     \midrule
     No fontanelle bulging & 3.62 & Neck tenderness & 2.44 & Bike injury & 2.01 \\
     Amnesia & 2.07 & Neck pain & 2.18 & Abdomen pain & 1.66 \\
     Pedestrian struck by vehicle  & 1.44 & Motor vehicle injury: other & 1.54 & Thoracic tenderness & 1.43 \\
     Headache & 1.39 & Hit by car & 1.47 & Hypotension & 1.23 \\
     Bike injury & 1.26 & Substantial injury: extremity & 1.35 & No abdomen pain & 0.98 \\
    \bottomrule
\end{tabular}
    \vspace{-5pt}
    \caption{Logistic regression coefficients for features that contribute to high $P({\ge}2 \textit{ yrs} \ | \ X)$ reflect  known medical knowledge. For example, features with large coefficients require verbal responses (e.g., \textit{Amnesia}, \textit{Headache}, \textit{Pain}), relate to activities not typical for the ${<}2$ \textit{yrs} group (\textit{Bike injury}), or are specific to older children, e.g., older children should have \textit{No fontanelle bulging}, as cranial soft spots typically close by 2 to 3 months after birth.}
    \label{tab:prop_model_variables}
\end{table*}

\paragraph{Interpreting the group membership model.}
In this clinical context, we begin by fitting several logistic regression and gradient-boosted decision tree group membership models to each of the training datasets to predict whether a patient is in the ${<}2$ \textit{yrs} or ${\ge}2$ \textit{yrs} group.
For the instance-weighted methods, we treat the choice of group membership model as a hyperparameter, and select the best model according to the downstream performance of the final decision rule on the validation set. 

\cref{tab:prop_model_variables} shows the coefficients of the most important features for each logistic regression group membership model when predicting whether a patient is in the ${\ge}2$ \textit{yrs} group.
The coefficients reflect existing medical expertise. For example, the presence of verbal response features (e.g., \textit{Amnesia}, \textit{Headache}) increases the probability of being in the ${\ge}2$ \textit{yrs} group, as does the presence of
activities not typical for the ${<}2$ \textit{yrs} group (e.g. \textit{Bike injury}).


\paragraph{Interpreting the outcome model.}

\cref{fig:model_ex_csi} shows the \methodabbrv~model on the CSI dataset, selected via cross-validation.
Outcome predictions for a group
are made by summing the predicted risk contribution (\textit{$\Delta$ Risk}) from the appropriate leaf of each tree in the group's fitted tree ensemble.\footnote{$\Delta$ Risk is not simply equivalent to the fraction of patients with CSI since (i) \methodabbrv~uses patients from both groups and (ii) each tree in FIGS fits the residuals of the others.}

The features used by each group are overlapping and reasonable, matching medical domain knowledge and partially matching previous work~\cite{leonard2019cervical}; e.g., features such as \textit{focal neuro signs}, \textit{neck pain}, and \textit{altered mental status} are all known to increase the risk of CSI.
Features unique to each group largely relate to the age cutoff; the ${<}2$ \textit{yrs} features only include those that clinicians can assess without asking the patient (e.g., \textit{substantial torso injury}), while two of the ${\ge}2$ \textit{yrs} features require verbal responses (\textit{neck pain}, \textit{head pain}).
\cref{sec:learned_models} shows fitted models for other datasets/methods;
the tree ensemble of \methodabbrv~allows it to adapt a succinct model to independent risk factors in the data whereas individual tree models (i.e., CART, TAO) are not flexible enough to model additive effects in the data.

\paragraph{Discussion.}
\methodabbrv~makes an important step towards interpretable modeling of heterogeneous data in the context of high-stakes clinical decision-making, with interesting avenues for future work. The fitted models show promise, but require external clinical validation before potential use. Our scope is limited to age-based splits in the clinical domain, but the behavior of \methodabbrv~with temporal, geographical, or demographic splits could be studied as well, on these or other datasets. Additionally, there are many methodological extensions to explore, such as data-driven identification of input data groups and schemes for feature weighting in addition to instance weighting. 

\section*{Acknowledgements}
We gratefully acknowledge partial support from NSF Grants DMS-1613002, 1953191, 2015341, IIS 1741340, the Center for Science of Information (CSoI), an NSF Sci- ence and Technology Center, under grant agreement CCF- 0939370, NSF grant 2023505 on Collaborative Research: Foundations of Data Science Institute (FODSI), the NSF and the Simons Foundation for the Collaboration on the Theoretical Foundations of Deep Learning through awards DMS-2031883 and 814639, and a grant from the Weill Neurohub.

\FloatBarrier
{
    \footnotesize
    \bibliography{main}
    \bibliographystyle{icml2022}
}

\newpage
\appendix
\onecolumn


\section{Fitted models}
\label{sec:learned_models}

\begin{figure}[h!]
    \centering
    \includegraphics[width=0.8\textwidth]{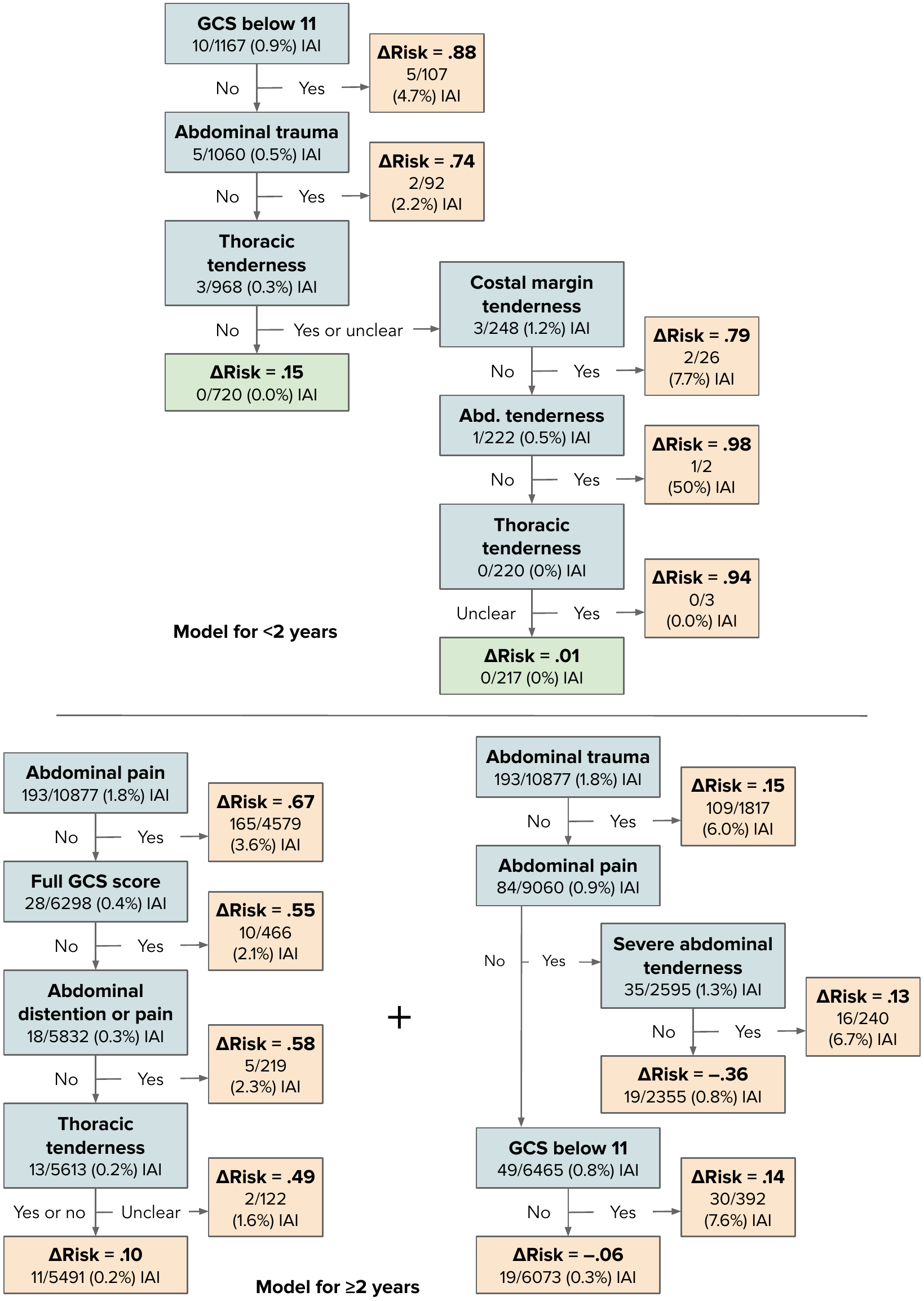}
    \caption{\methodabbrv~model fitted to the IAI dataset. Note that the younger group only uses \textit{tenderness}, which can evaluated without verbal input from the patient, whereas the older group uses \textit{pain}, which requires a verbal response. Achieves 95.1\% sensitivity and 50.8\% specificity (training).}
    \label{fig:model_ex_iai}
    \vspace{20pt}
\end{figure}

\begin{figure}[H]
    \centering
    \vspace{50pt}
    \includegraphics[width=0.95\textwidth]{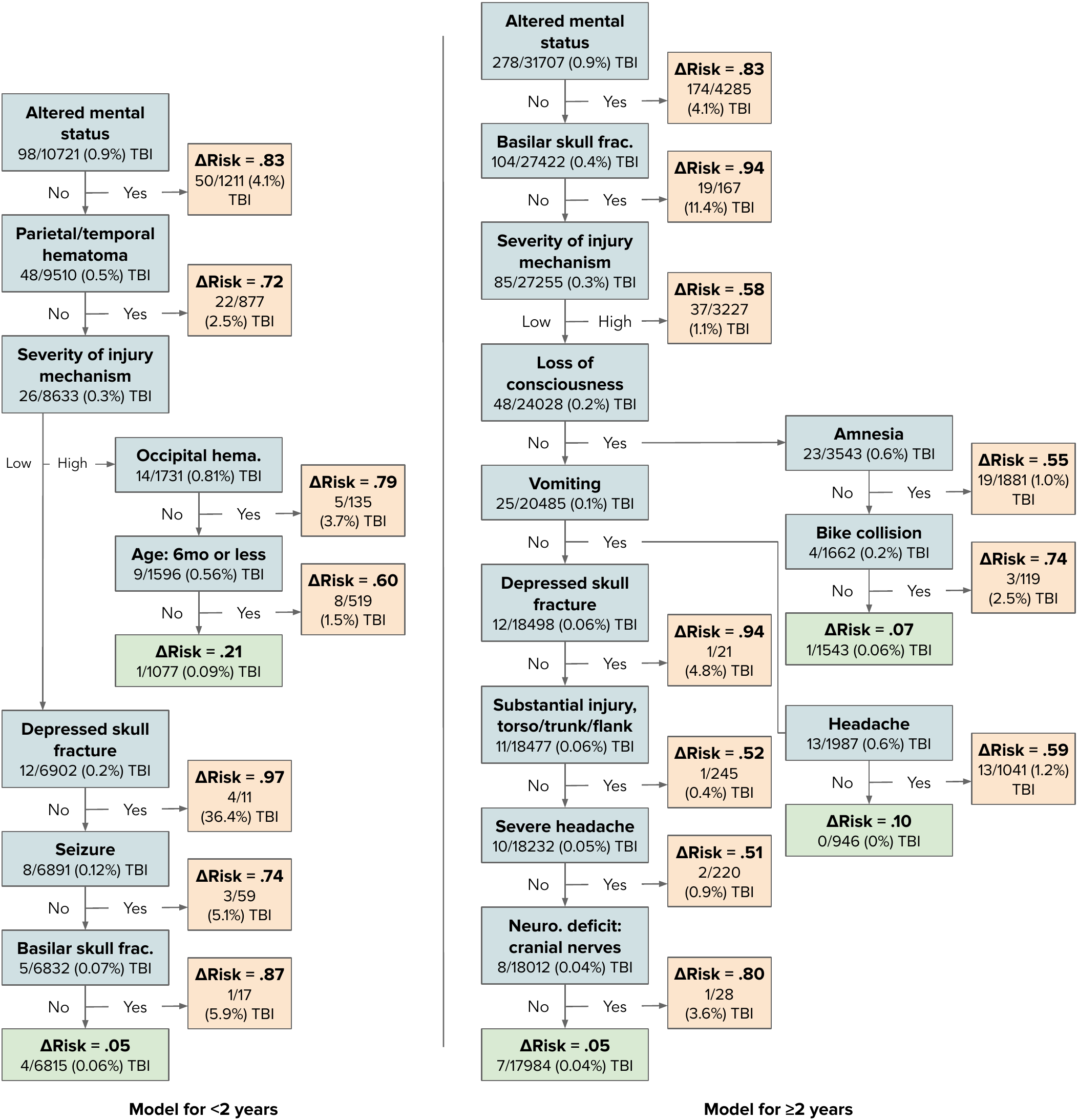}
    \caption{\methodabbrv~model fitted to the TBI dataset. Interestingly, in this case \methodabbrv~learns only a single tree for each group. Note that the model for the older group utilizes the \textit{Headache} and \textit{Severe Headache} features, which require a verbal response. Achieves 97.1\% sensitivity and 58.9\% specificity (training).}
    \label{fig:model_ex_tbi}
\end{figure}

\begin{figure}[H]
    \centering
    \vspace{50pt}
    \includegraphics[width=0.95\textwidth]{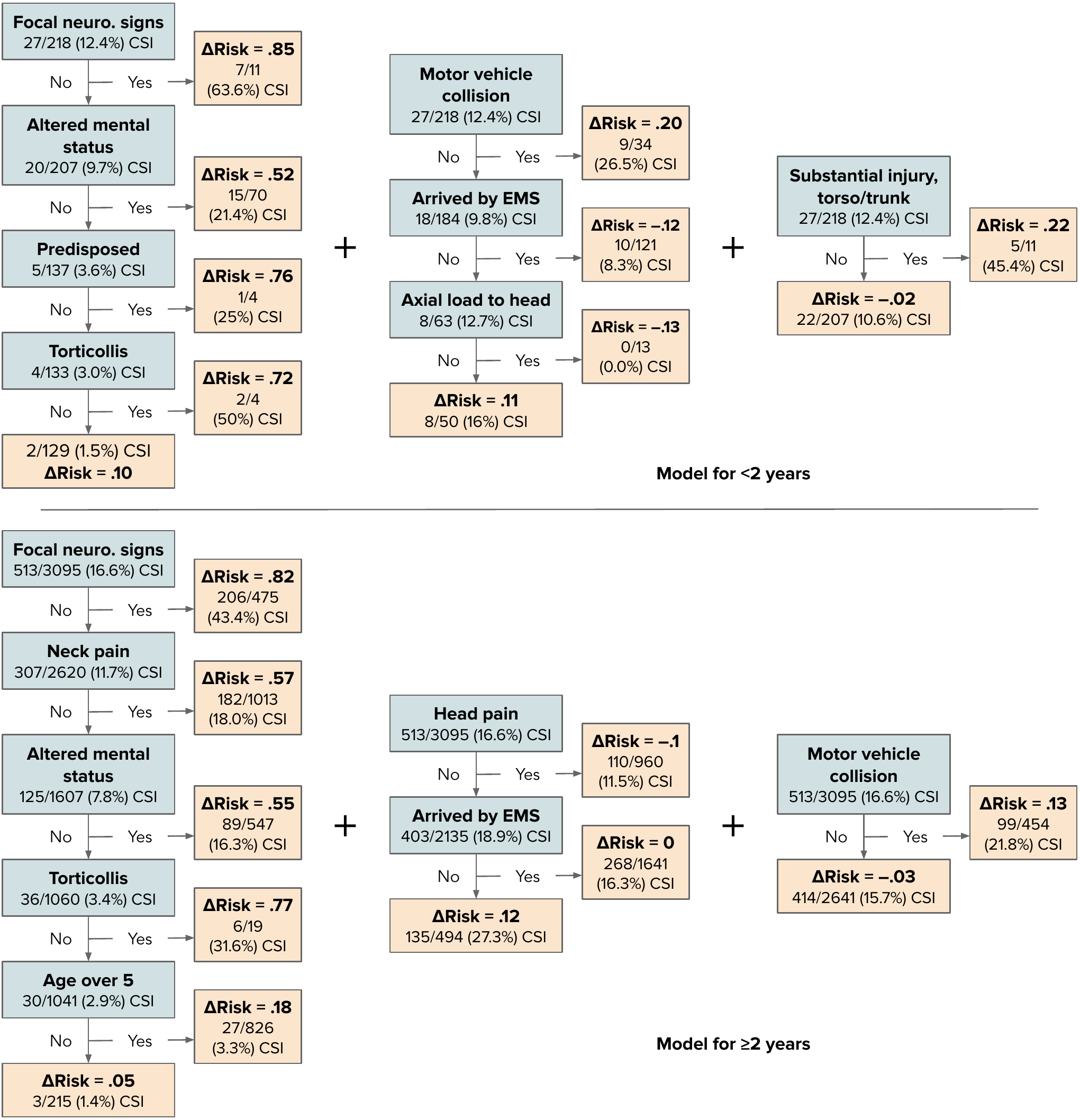}
    \caption{\methodabbrv~model fitted to the CSI dataset (duplicated from the main text~\cref{fig:model_ex_csi}). Achieves 97.0\% sensitivity and 33.9\% specificity (training). The left tree for ${<}2$ \textit{yrs} gives large $\Delta$ Risk to active features, and on its own provides sensitivity of 99\%.
    Counterintuitively, the middle tree assigns $\Delta$ Risk $<0$ for patients arriving by ambulance (\textit{EMS}) or with head injuries that affect the spine (\textit{axial load}).
    However, adding this second tree results in boosted specificity (increase of 8.7\%) with a tiny reduction in sensitivity (decrease of 0.4\%), indicating that \methodabbrv~adaptively tunes the sensitivity-specificity tradeoff.}
    \label{fig:model_ex_csi_rep}
\end{figure}


\section{Simulation}
\label{sec:simulations}

\begin{figure}[H]
    \centering  
    \includegraphics[width=9cm]{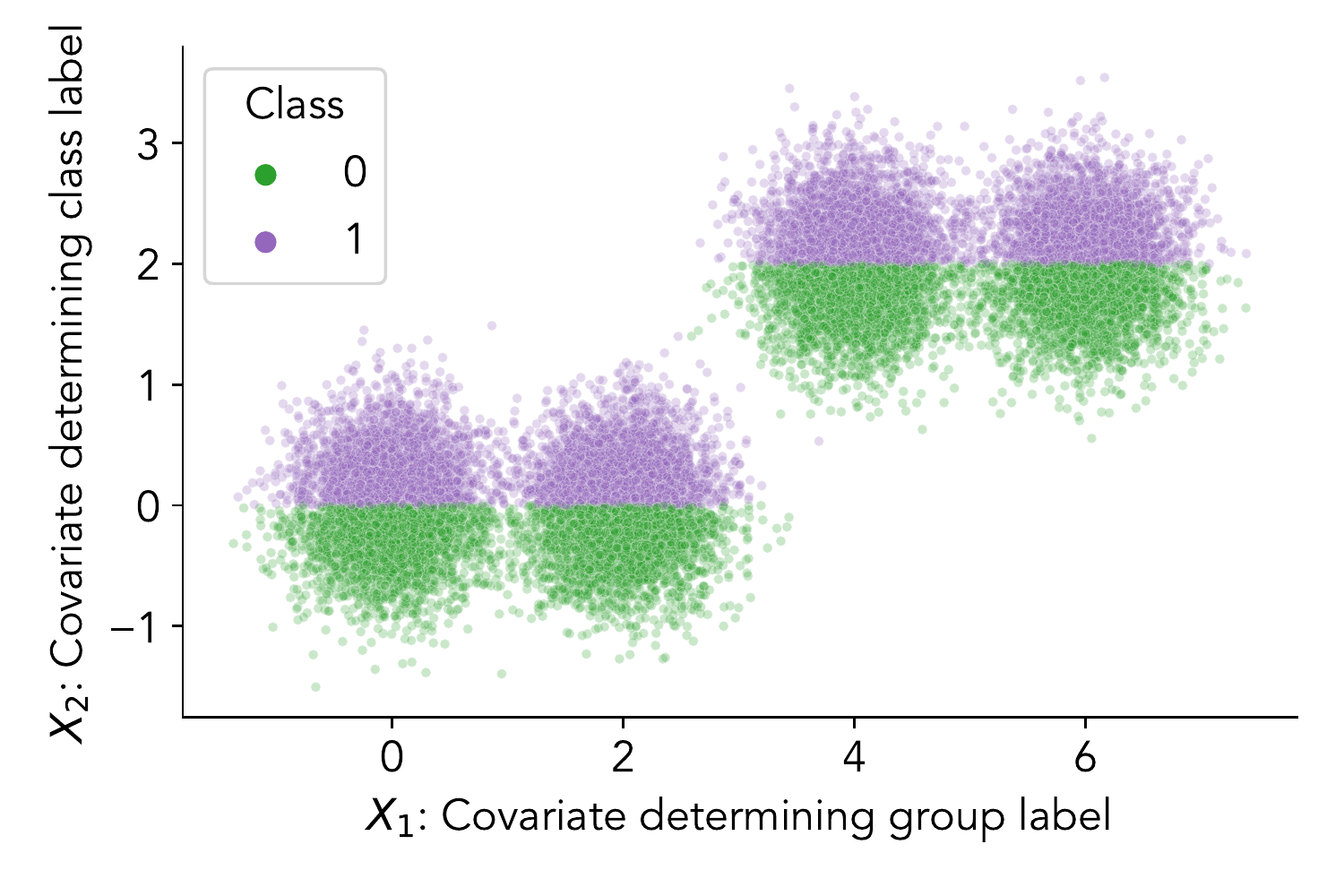}
    \caption{Visualization of the data-generating process for the simulation. Each cluster represents a group for \methodabbrv. The two clusters on the left and two clusters on the right share a prediction rule, presenting a simple case where sharing data between groups can help performance. Noise variables are not pictured, and the variances of $X_1$ and $X_2$ are reduced for a clearer visualization.}
    \label{fig:simdata}
\end{figure}

In addition our evaluations on clinical datasets, we evaluate \methodabbrv~under a simple simulation involving heterogeneous data. 
The data-generating process is multivariate Gaussian with four clusters and two meta-clusters which share the same relationship between $X$ and $Y$, visualized in \cref{fig:simdata}. There are two variables of interest, $X_1$ and $X_2$, and 10 noise variables. Each cluster is centered at a different value of $X_1$; the first meta-cluster consists of the clusters centered at $X_1 = 0$ and $X_1 = 2$, which share the relationship $Y = X_2 > 0$, while the second consists of the clusters centered at $X_1 = 4$ and $X_1 = 6$, which share the relationship $Y = X_2 > 2$. $X_1$ and $X_2$ have variance 1 and all noise variables have variance 2; additionally, zero-mean noise with variance 2 is added to $X_1$ and $X_2$.

The four clusters are then treated as four groups, to which separate models are fitted. If the intuition behind \methodabbrv~is correct, \methodabbrv~should assign relatively higher probabilities to points that are within a given cluster's meta-cluster, and relatively lower probabilities to points in the other meta-cluster. In comparison to fitting completely separate models, this should increase the amount of data available for learning the two rules, thereby counteracting noise and resulting in better performance. On the other hand, if one model is fit to all of the data, we expect the lack of group-awareness to hurt performance (i.e. the crucial split at $X_1 = 3$ may be missed since it does not significantly reduce entropy).  Our evaluation suggests that this is the case; as shown in \cref{tab:simresults}, G-CART and \methodabbrv~significantly outperform baseline methods. 

We do not perform any hyperparameter selection; we fix the maximum number of tree splits to be 1 for the probability-weighted models and \textit{-SEP} models, and 4 for the models fit to all the data. The rationale for this is that 3 splits are sufficient to ideally model the entire data-generating process (splits at $X_1 = 3$, $X_2 = 0$, and $X_2 = 2$) and 1 split is sufficient for each cluster. Note that when only one split is used, G-CART and \methodabbrv~are the same algorithm. Logistic regression is used to fit the group membership model.

\begin{table}[H]
    \centering
    \small
    \begin{tabular}{lcccc}
    \toprule
    {} & ROC AUC & APS & Accuracy & F1 \\
    \midrule
    TAO & .376 (.07) & .498 (.04) & 59.0 (.02) & 58.0 (.04) \\
    TAO-SEP  & .475 (.04) & .573 (.03) & 58.3 (.02) & 60.4 (.03) \\
    CART & .370 (.07) & .495 (.04) & 56.5 (.02) & 54.7 (.03) \\
    CART-SEP & .475 (.04) & .573 (.03) & 58.3 (.02) & 60.4 (.03) \\
    FIGS & .470 (.04) & .539 (.04) & 58.5 (.02) & 55.5 (.03) \\
    FIGS-SEP & .475 (.04) & .573 (.03) & 58.3 (.02) & 60.4 (.03) \\
    \textbf{G-CART / \methodabbrv} & \textbf{.550} (.03) & \textbf{.644} (.03) & \textbf{65.8} (.03)  & \textbf{63.9} (.04) \\
    \bottomrule
\end{tabular}
        \caption{Unlike the clinical datasets, the simulation data is class-balanced and lacks a medical context, so we report area under the ROC curve, average precision score, accuracy, and F1 score instead of specificity metrics. Because only one split per cluster is computed for G-CART and \methodabbrv~they reduce to the exact same algorithm, so their results are shown together.}
    \label{tab:simresults}
\end{table}

\section{Hyperparameter selection}
\label{sec:hyperparams}

\begin{table}[H]
    \centering
    \footnotesize
    \begin{tabular}{lrrrrrr}
\toprule
{} & \multicolumn{3}{c}{${<}2$ \textit{yrs} group} &
        \multicolumn{3}{c}{${\ge}2$ \textit{yrs} group} \\
    \cmidrule(lr){2-4}
    \cmidrule(lr){5-7}
Maximum tree splits: &           8 &          12 &          16 &           8 &          12 &          16 \\
\midrule
TAO (1 iter)                 &  \textbf{15.1} (6.7) &  15.1 (6.7) &  14.4 (6.1) &  \textbf{14.1} (7.8) &  14.1 (7.8) &   8.9 (5.9) \\
TAO (5 iter)                &  \textbf{14.4} (6.1) &   0.0 (0.0) &   0.0 (0.0) &   \textbf{8.9} (5.9) &   3.1 (0.9) &   1.5 (0.7) \\
CART-SEP                    &  \textbf{15.1} (6.7) &  14.4 (6.1) &   0.0 (0.0) &  \textbf{14.0} (7.8) &   8.9 (5.9) &   3.1 (0.9) \\
FIGS-SEP                    &  \textbf{13.7} (5.9) &   0.0 (0.0) &   0.0 (0.0) &  \textbf{23.1} (8.8) &  13.0 (7.4) &   7.8 (5.6) \\
G-CART w/ LR ($C = 2.8$)       &   \textbf{7.9} (6.7) &   3.1 (2.1) &   3.5 (1.7) &  19.0 (8.8) &  \textbf{21.8} (8.4) &   2.1 (0.6) \\
G-CART w/ LR ($C = 0.1$)       &  \textbf{20.4} (8.6) &   8.3 (6.6) &  10.1 (6.7) &  12.7 (7.6) &  \textbf{14.9} (7.1) &   3.6 (0.9) \\
G-CART w/ GB ($N = 100$)       &  \textbf{19.8} (8.3) &   7.2 (6.3) &   7.6 (6.1) &  13.3 (8.0) &  \textbf{21.4} (8.5) &   9.0 (5.6) \\
G-CART w/ GB ($N = 50$)        &  \textbf{26.8} (9.7) &   8.1 (6.3) &   8.4 (6.1) &  13.3 (8.0) &  \textbf{21.4} (8.5) &   9.7 (5.6) \\
\methodabbrv~w/ LR ($C = 2.8$) &  \textbf{14.9} (8.5) &   7.5 (5.4) &   8.1 (6.9) &  41.0 (8.7) &  \textbf{48.1} (8.2) &  35.6 (8.9) \\
\methodabbrv~w/ LR ($C = 0.1$) &  \textbf{31.0} (9.4) &  23.1 (9.1) &  25.9 (9.7) &  46.9 (8.4) &  \textbf{48.2} (8.4) &  33.7 (8.9) \\
\methodabbrv~w/ GB ($N = 100$) &  \textbf{24.5} (8.6) &  24.0 (9.3) &  21.2 (8.7) &  \textbf{47.5} (8.5) &  47.5 (8.2) &  27.9 (8.6) \\
\methodabbrv~w/ GB ($N = 50$)  &  \textbf{32.1} (9.6) &  18.3 (8.2) &  12.7 (6.9) &  47.5 (8.5) &  \textbf{53.2} (7.3) &  28.4 (8.3) \\
\bottomrule
\vspace{1pt}
\end{tabular}

(a)

\vspace{10pt}

\begin{tabular}{lrrrr}
\toprule
Group membership model:  & LR ($C = 2.8$) & LR ($C = 0.1$) & GB ($N = 100$) & GB ($N = 50$) \\
\midrule
G-CART (${<}2$ \textit{yrs}, ${\ge}2$ \textit{yrs} models combined) &   \textbf{27.8} (6.0) &   21.5 (5.9) &     19.0 (5.7) &    27.1 (6.5) \\
\methodabbrv~(${<}2$ \textit{yrs}, ${\ge}2$ \textit{yrs} models combined) &   51.3 (5.8) &   54.5 (6.2) &     \textbf{57.4} (5.6) &    44.6 (7.4) \\
\bottomrule
\vspace{1pt}
\end{tabular}

(b)
        \caption{Hyperparameter selection table for the TBI dataset; the metric shown is specificity at 94\% sensitivity on the validation set, with corresponding standard error in parentheses. First, the best-performing maximum of tree splits is selected for each method or combination of method and membership model (a). This is done separately for each data group. Next, the best membership model is selected for G-CART and \methodabbrv~using the overall performance of the best models from (a) across both data groups (b). The two-stage validation process ensures that the ${<}2$ \textit{yrs} and ${\ge}2$ \textit{yrs} groups use the same group membership probabilities, which we have found leads to better performance than allowing them to use different membership models. Metrics shown are averages across the 10 validation sets, but hyperparameter selection was done independently for each of the 10 data splits.}
    \label{tab:cv}
\end{table}

\paragraph{Data splitting} We use 10 random training/validation/test splits for each dataset, performing hyperparameter selection separately on each. There are two reasons we choose not to use a fixed test set. First, the small number of positive instances in our datasets makes our primary metrics (specificity at high sensitivity levels) noisy, so averaging across multiple splits makes the results more stable. Second, the works that introduced the TBI, IAI, and CSI datasets did not publish their test sets, as it is not as common to do so in the medical field as it is in machine learning, making the choice of test set unclear. For TBI and CSI, we simply use the random seeds 0 through 10. For IAI, some filtering of seeds is required due to the low number of positive examples; we reject seeds that do not allocate positive examples evenly enough between each split (a ratio of negative to positive outcomes over 200 in any split).

\paragraph{Class weights} Due to the importance of achieving high sensitivity, we upweight positive instances in the loss by the inverse proportion of positive instances in the dataset. This results in class weights of about 7:1 for CSI, 112:1 for TBI, and 60:1 for IAI. These weights are fixed for all methods.

\paragraph{Hyperparameter settings} Due to the relatively small number of positive examples in all datasets, we keep the hyperparameter search space small to avoid overfitting. We vary the maximum number of tree splits from 8 to 16 for all methods and the maximum number of update iterations from 1 to 5 for TAO. The options of group membership model are logistic regression with L2 regularization and gradient-boosted trees \citep{friedman2001greedy}. For both models, we simply include two hyperparameter settings: a less-regularized version and a more-regularized version, by varying the inverse regularization strength ($C$) for logistic regression and the number of trees ($N$) for gradient-boosted trees. We initially experimented with random forests and CART, but found them to lead to poor downstream performance. Random forests tended to separate the groups too well in terms of estimated probabilities, leading to little information sharing between groups, while CART did not provide unique enough membership probabilities, since CART probability estimates are simply within-node class proportions.

\paragraph{Validation metrics} We use the highest specificity achieved when sensitivity is at or above 94\% as the metric for validation. If this metric is tied between different hyperparameter settings of the same model, specificity at 90\% sensitivity is used as the tiebreaker. For the IAI dataset, only specificity at 90\% sensitivity is used, since the relatively small number of positive examples makes high sensitivity metrics noisier than usual. If there is still a tie at 90\% sensitivity, the smaller model in terms of number of tree splits is chosen.

\paragraph{Validation of group membership model} Hyperparameter selection for \methodabbrv~and G-CART is done in two stages due to the need to select the best group membership model. First, the best-performing maximum of tree splits is selected for each combination of method and membership model. This is done separately for each data group. Next, the best membership model is selected using the overall performance of the best models across both data groups. The two-stage validation process ensures that the ${<}2$ \textit{yrs} and ${\ge}2$ \textit{yrs} groups use the same group membership probabilities, which we have found performs better than allowing different sub-models of \methodabbrv~to use different membership models.

\section{Data preprocessing details}
\label{sec:data_preprocessing}

\paragraph{Traumatic brain injury (TBI)}

To screen patients, we follow the inclusion and exclusion criteria from \citet{kuppermann2009identification}, which exclude patients with Glasgow Coma Scale (GCS) scores under 14 or no signs or symptoms of head trauma, among other disqualifying factors. No patients were dropped due to missing values: the majority of patients have about 1\% of features missing, and are at maximum still under 20\%. We utilize the same set of features as \citet{kuppermann2009identification}.

Our strategy for imputing missing values differed between features according to clinical guidance. For features that are unlikely to be left unrecorded if present, such as paralysis, missing values were assumed to be negative. For other features that could be unnoticed by clinicians or guardians, such as loss of consciousness, missing values are assumed to be positive. For features that did not fit into either of these groups or were numeric, missing values are imputed with the median.

\paragraph{Cervical spine injury (CSI)}

\citet{leonard2019cervical} engineered a set of 22 expert features from 609 raw features; we utilize this set but add back features that provide information on the following:
\begin{itemize}
    \vspace{-5pt}
    \setlength\itemsep{0.1em}
    \item Patient position after injury
    \item Clinical intervention received by patients prior to arrival (immobilization, intubation)
    \item Pain and tenderness of the head, face, torso/trunk, and extremities
    \item Age and gender
    \item Whether the patient arrived by emergency medical service (EMS)
    \vspace{-5pt}
\end{itemize}

We follow the same imputation strategy described in the TBI subsection above. Features that are assumed to be negative if missing include focal neurological findings, motor vehicle collision, and torticollis, while the only feature assumed to be positive if missing is loss of consciousness.

\paragraph{Intra-abdominal injury (IAI) }

We follow the data preprocessing steps described in \citet{holmes2013identification} and \cite{kornblith2022predictability}. In particular, all features of which at least 5\% of values are missing are removed, and variables that exhibit insufficient interrater agreement (lower bound of 95\% CI under 0.4) are removed.
The remaining missing values are imputed with the median.
In addition to the 18 original variables, we engineered three additional features:
\begin{itemize}
    \vspace{-5pt}
    \setlength\itemsep{0.1em}
    \item \textit{Full GCS score}: True when GCS is equal to     the maximum score of 15
    \item \textit{Abd. Distention or abd. pain}: Either         abdominal distention or abdominal pain 
    \item \textit{Abd. trauma or seatbelt sign}: Either         abdominal trauma or seatbelt sign
    \vspace{-5pt}
\end{itemize}

\paragraph{Data for predicting group membership probabilities}

The data preprocessing steps for the group membership models in the first step of \methodabbrv~are identical to that above, except that missing values are not imputed at all for categorical features, such that ``missing", or NaN, is allowed as one of the feature labels in the data. We find that this results in more accurate group membership probabilities, since for some features, such as those requiring a verbal response, missing values are predictive of age group.

Unprocessed data is available at \url{https://pecarn.org/datasets/} and clean data is available on github at 
\url{https://github.com/csinva/imodels-data} (easily accessibly through the imodels package~\cite{singh2021imodels}).

\begin{table}[H]
    \centering
    \footnotesize
    
\begin{tabular}{cc}

\begin{tabular}{lrr}
\toprule
\multicolumn{3}{c}{Traumatic brain injury} \\
\midrule
                Feature Name &  \% Missing & \% Nonzero \\
\midrule
    Altered Mental Status &              0.74 &     12.95 \\
    Altered Mental Status: Agitated &             87.05 &      1.79 \\
    Altered Mental Status: Other &             87.05 &      1.82 \\
    Altered Mental Status: Repetitive &             87.05 &      1.04 \\
    Altered Mental Status: Sleepy &             87.05 &      6.67 \\
    Altered Mental Status: Slow to respond &             87.05 &      3.22 \\
    Acting normally per parents &              7.09 &     85.38 \\
    Age (months) &              0.00 &       N/A \\
    Verbal amnesia &             38.41 &     10.45 \\
    Trauma above clavicles &              0.30 &     64.38 \\
    Trauma above clavicles: Face &             35.92 &     29.99 \\
    Trauma above clavicles: Scalp-frontal &             35.92 &     20.48 \\
    Trauma above clavicles: Neck &             35.92 &      1.38 \\
    Trauma above clavicles: Scalp-occipital &             35.92 &      9.62 \\
    Trauma above clavicles: Scalp-parietal &             35.92 &      7.79 \\
    Trauma above clavicles: Scalp-temporal &             35.92 &      3.39 \\
    Drugs suspected &              4.19 &      0.87 \\
    Fontanelle bulging &              0.37 &      0.06 \\
    Sex &              0.01 &       N/A \\
    Headache severity &              2.38 &       N/A \\
    Headache start time &              3.09 &       N/A \\
    Headache &             32.76 &     29.94 \\
    Hematoma &              0.69 &     39.42 \\
    Hematoma location &              0.47 &       N/A \\
    Hematoma size &              1.67 &       N/A \\
    Severity of injury mechanism &              0.74 &       N/A \\
    Injury mechanism &              0.67 &       N/A \\
    Intubated &              0.73 &      0.01 \\
    Loss of consciousness &              4.05 &     10.37 \\
    Length of loss of consciousness&              5.39 &       N/A \\
    Neurological deficit &              0.85 &       1.3 \\
    Neurological deficit: Cranial &             98.70 &      0.18 \\
    Neurological deficit: Motor &             98.70 &      0.28 \\
    Neurological deficit: Other &             98.70 &      0.71 \\
    Neurological deficit: Reflex &             98.70 &      0.03 \\
    Neurological deficit: Sensory &             98.70 &      0.26 \\
    Other substantial injury  &              0.43 &     10.07 \\
    Other substantial injury: Abdomen &             89.93 &      1.25 \\
    Other substantial injury: Cervical spine &             89.93 &      1.37 \\
    Other substantial injury: Cut &             89.93 &      0.12 \\
    Other substantial injury: Extremity &             89.93 &      5.49 \\
    Other substantial injury: Flank &             89.93 &      1.56 \\
    Other substantial injury: Other &             89.93 &      1.65 \\
    Other substantial injury: Pelvis &             89.93 &      0.44 \\
    Paralyzed &              0.75 &      0.01 \\
    Basilar skull fracture &              0.99 &      0.68 \\
    Basilar skull fracture: Hemotympanum &             99.32 &      0.35 \\
    Basilar skull fracture: CSF otorrhea &             99.32 &      0.04 \\
    Basilar skull fracture: Periorbital &             99.32 &      0.19 \\
    Basilar skull fracture: Retroauricular &             99.32 &      0.08 \\
    Basilar skull fracture: CSF rhinorrhea & 99.32 &      0.03 \\
    Skull fracture: Palpable &              0.24 &      0.38 \\
    Skull fracture: Palpable and depressed &             99.69 &      0.18 \\
    Sedated &              0.76 &      0.08 \\
    Seizure &              1.70 &      1.17 \\
    Length of seizure &              0.18 &       N/A \\
    Time of seizure &              0.12 &       N/A \\
    Vomiting &              0.71 &      13.1 \\
    Time of last vomit &             89.04 &       N/A \\
\end{tabular}
 &

\begin{tabular}{lrr}

Number of times vomited &              0.60 &       N/A \\
Vomit start time &              0.87 &       N/A \\
\midrule
\multicolumn{3}{c}{Intra-abdominal injury} \\
\midrule
    Abdominal distention &              4.38 &       2.3 \\
    Abdominal distention or pain &              0.00 &      4.93 \\
    Degree of abdominal tenderness &             70.13 &       N/A \\
    Abdominal trauma &              0.56 &     15.48 \\
    Abdominal trauma or seat belt sign & 0.00 &      16.3 \\
    Abdomen pain &             15.38 &     30.06 \\
    Age (years) &              0.00 &       N/A \\
    Costal margin tenderness &              0.00 &     11.33 \\
    Decreased breath sound &              1.93 &      2.13 \\
    Distracting pain &              7.38 &     23.29 \\
    Glasgow Coma Scale (GCS) score &              0.00 &       N/A \\
    Full GCS score &              0.00 &     86.21 \\
    Hypotension &              0.00 &      1.44 \\
    Left costal margin tenderness &              0.00 &       N/A \\
    Method of injury &              3.95 &       N/A \\
    Right costal margin tenderness &              0.00 &       N/A \\
    Seat belt sign &              3.30 &      4.93 \\
    Sex &              0.00 &       N/A \\
    Thoracic tenderness &              9.99 &     15.96 \\
    Thoracic trauma &              0.63 &     16.95 \\
    Vomiting &              3.92 &      9.57 \\
\midrule
\multicolumn{3}{c}{Cervical spine injury} \\
\midrule
                  Age (years) &              0.00 &       N/A \\
        Altered mental status &              2.05 &     24.72 \\
             Axial load to head &              0.00 &      24.0 \\
               Clotheslining &              3.38 &      0.94 \\
         Focal neurological findings &              9.84 &     14.67 \\
              Method of injury: Diving &              0.03 &       1.3 \\
                Method of injury: Fall &              2.44 &      3.83 \\
             Method of injury: Hanging &              0.03 &      0.15 \\
            Method of injury: Hit by car &              0.03 &     15.09 \\
                 Method of injury: Auto collision &              7.73 &     14.73 \\
             Method of injury: Other auto &              0.03 &      3.11 \\
                Arrived by EMS &              0.00 &     77.24 \\
                         Loss of consciousness  &              8.03 &     42.68 \\
                   Neck pain &              5.25 &     38.42 \\
       Posterior midline neck tenderness &              2.57 &     29.88 \\
                    Patient position on arrival &             61.52 &       N/A \\
                 Predisposed &              0.00 &      0.66 \\
              Pain: Extremity &             18.35 &     25.87 \\
             Pain: Face &             18.35 &      7.58 \\
             Pain: Head &             18.35 &     29.04 \\
       Pain: Torso/trunk &             18.35 &     28.95 \\
                Tenderness: Extremity &             20.37 &     15.15 \\
               Tenderness: Face &             20.37 &      3.83 \\
               Tenderness: Head &             20.37 &      7.79 \\
         Tenderness: Torso/trunk &             20.37 &     25.87 \\
                  Substantial injury: Extremity &              1.03 &     10.87 \\
                 Substantial injury: Face &              1.06 &      5.67 \\
                 Substantial injury: Head &              1.00 &     15.88 \\
           Substantial injury: Torso/trunk &              1.03 &       7.3 \\
                 Neck tenderness &              2.48 &      39.3 \\
                Torticollis &              7.03 &      5.77 \\
                  Ambulatory &              5.77 &     21.46 \\
                Axial load to top of head &              0.00 &      2.35 \\
                Sex &              0.00 &       N/A \\
\bottomrule
\end{tabular}
\end{tabular}

        \caption{Final features used for fitting the \textit{outcome} models. Features include information about patient history (i.e. \textit{mechanism of injury}), physical examination (i.e. \textit{Abdominal trauma}), and mental condition (i.e. \textit{Altered mental status}). Percentage of nonzero values is marked \textit{N/A} for non-binary features.}
    \label{tab:features}
\end{table}

\section{Extended results}

\begin{table}[H]
    \centering
    \footnotesize
    \begin{tabular}{lrrrrrrrrr}
    \toprule
    {} & \multicolumn{6}{c}{Traumatic brain injury} &
        \multicolumn{3}{c}{Cervical spine injury} \\
     \cmidrule(lr){2-7}
     \cmidrule(lr){8-10}
     {} & 92\% & 94\% & 96\% & 98\% & ROC AUC & F1
     & 92\% & 94\% & 96\% \\
     \midrule
    TAO & 6.2 (5.9) & 6.2 (5.9) & 0.4 (0.4) & 0.4 (0.4) &  .294 (.05) &  5.2 (.00) 
        & 41.5 (0.9) & 21.2 (6.6) & 0.2 (0.2)   \\
     TAO-SEP & 26.7 (6.4) & 13.9 (5.4) & 10.4 (5.5) & 2.4 (1.5) &  .748 (.02) &  \textbf{5.8} (.00)
        & 32.5 (4.9) & 7.0 (1.6) & 5.4 (0.7)  \\
    CART & 20.9 (8.8) & 14.8 (7.6) & 7.8 (5.8) & 2.1 (0.6) &  .702 (.06) &  5.7 (.00)
        & 38.6 (3.6) & 13.7 (5.7) & 1.5 (0.6)  \\
     CART-SEP & 26.6 (6.4) &  13.8 (5.4) &  10.3 (5.5) &  2.4 (1.5) &  .753 (.02) &  5.6 (.00)
        & 32.1 (5.1) & 7.8 (1.5) & 5.4 (0.7)  \\
    G-CART & 15.5 (5.5) &  13.5 (5.7) &   6.4 (2.2) &  3.0 (1.5) &  \textbf{.758} (.01) &  5.5 (.00)
        &  38.5 (3.4) &  15.2 (4.8) & 4.9 (1.0) \\
     FIGS & 23.8 (9.0) &  18.2 (8.5) &  12.1 (7.3) &  0.4 (0.3) &  .380 (.07) &  4.8 (.00)
        & 39.1 (3.0) & 33.8 (2.4) & 24.2 (3.2)   \\
     FIGS-SEP & 39.9 (7.9) & 19.7 (6.8) &  \textbf{17.5} (7.0) &  2.6 (1.6) &  .619 (.05) &  5.1 (.00)
        & 38.7 (1.6) & 33.1 (2.0) & 20.1 (2.6)  \\
     \textbf{\methodabbrv} & \textbf{42.0} (6.6) &  \textbf{23.0} (7.8) & 14.7 (6.5) &  \textbf{6.4} (2.8) &  .696 (.04) &  4.7 (.00)
        & \textbf{42.2} (1.3) & \textbf{36.2} (2.3) & \textbf{28.4} (3.8)  \\
    \midrule
\end{tabular}


\begin{tabular}{lrrrrrrrrr}
    {} & \multicolumn{3}{c}{Cervical spine injury (cont.)}
       & \multicolumn{6}{c}{Intra-abdominal injury}\\
     \cmidrule(lr){2-4}
     \cmidrule(lr){5-10}
     {} & 98\% & ROC AUC & F1 & 92\% & 94\% & 96\% & 98\% & ROC AUC & F1\\
     \midrule
     TAO & 0.2 (0.2) &  .422 (.04) &  44.5 (.01)
        & 0.2 (0.2) & 0.2 (0.2) & 0.0 (0.0) & 0.0 (0.0) & .372 (.04) &  \textbf{13.9} (.01) \\
     TAO-SEP & 2.5 (1.0) &  .702 (.01) &  44.4 (.01)
        & 12.1 (1.7) & 8.5 (2.0) & 2.0 (1.3) & 0.0 (0.0) & .675 (.01) &  12.9 (.00) \\
     CART & 1.1 (0.4) &  .617 (.06) &  \textbf{45.8} (.01)
        & 11.8 (5.0) & 2.7 (1.0) & 1.6 (0.5) & 1.4 (0.5) &  .688 (.06) &  13.4 (.00) \\
     CART-SEP & 2.5 (1.0) &  .707 (.00) &  44.2 (.01)
        & 11.0 (1.6) & 9.3 (1.8) & 2.8 (1.4) & 0.0 (0.0) &  .688 (.01) &  13.0 (.01) \\
     G-CART & 3.9 (1.1)  &  \textbf{.751} (.01) &  45.2 (.01)
        & 11.7 (1.3) &  10.1 (1.6) & 3.8 (1.3) &  0.7 (0.4) &  \textbf{.732} (.02) &  12.5 (.01) \\
     FIGS & \textbf{16.7} (3.9) &  .664 (.03) &  43.0 (.01)
        & \textbf{32.1} (5.5) & 13.7 (6.0) & 1.4 (0.8) & 0.0 (0.0) &  .541 (.04) &   9.4 (.01)\\
     FIGS-SEP & 3.9 (2.2) &  .643 (.02) &  41.4 (.01)
        & 18.8 (4.4) &  9.2 (2.2) & 2.6 (1.7) &  0.9 (0.8) &  .653 (.02) &   8.0 (.00)\\
     \textbf{\methodabbrv} & 15.7 (3.9) &  .700 (.01) &  42.6 (.01)
        & 29.7 (6.9) &  \textbf{18.8} (6.6) & \textbf{11.7} (5.1) & \textbf{3.0} (1.3) &  .671 (.03) &   9.1 (.01) \\
    \bottomrule
\end{tabular}
        \caption{Test set prediction results averaged over 10 random data splits, with corresponding standard error in parentheses. Values in columns labeled with a sensitivity percentage (e.g. 92\%) are best specificity achieved at the given level of sensitivity or greater. \methodabbrv~provides the best performance overall in the high-sensitivity regime. G-CART attains the best ROC curves, while TAO is strongest in terms of F1 score. 
        }
    \label{tab:results_extended}
\end{table}

We include the results from above with their standard errors, as well as additional metrics (Area under the ROC Curve and F1 score) for each dataset.




\end{document}